\documentclass{article} % For LaTeX2e
\usepackage{iclr2023_conference_tinypaper,times}

\usepackage{amsmath,amsfonts,bm}

\def\eqref#1{equation~\ref{#1}}
\def\1{\bm{1}}

\DeclareMathAlphabet{\mathsfit}{\encodingdefault}{\sfdefault}{m}{sl}
\SetMathAlphabet{\mathsfit}{bold}{\encodingdefault}{\sfdefault}{bx}{n}

\usepackage{hyperref}
\usepackage{url}
\usepackage{lipsum}
\usepackage{booktabs}
\usepackage{wrapfig}
\usepackage{graphicx}

\title{The Obscure Limitation of Modular\\Multilingual Language Models}

\iclrfinalcopy 
\author{Muhammad Farid Adilazuarda$\star^1$, Samuel Cahyawijaya$\star^2$, Ayu Purwarianti$^1$ \\
$^1$Institut Teknologi Bandung \quad $^2$ HKUST\\
\texttt{faridlazuarda@gmail.com, scahyawijaya@connect.ust.hk}
}

\begin{document}

\maketitle

\begin{abstract}

We expose the limitation of modular multilingual language models (MLMs) in multilingual inference scenarios with unknown languages. Existing evaluations of modular MLMs exclude the involvement of language identification (LID) modules, which obscures the performance of real-case multilingual scenarios of modular MLMs. In this work, we showcase the effect of adding LID on the multilingual evaluation of modular MLMs and provide discussions for closing the performance gap of caused by the pipelined approach of LID and modular MLMs.

\end{abstract}

\begin{wrapfigure}{R}{0.33\textwidth}
\vspace{-3.8em}
\centering
\includegraphics[width=0.33\textwidth]{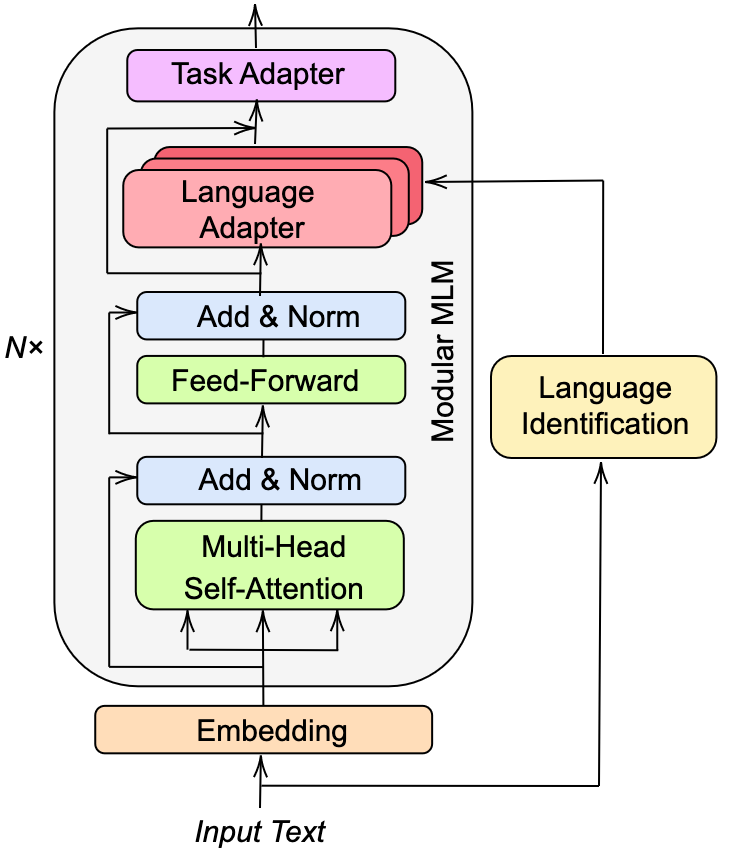}
\caption{Modular MLMs incorporate language-specific adapters to learn new languages. This renders them language-dependent and reliant on external LID for inference.}
\label{fig:overview}
\end{wrapfigure}

\section{Introduction}

Multilingual language models (MLMs) suffer from the capacity limitation problem known as the \textbf{curse of multilinguality}, which penalizes the efficiency of MLMs, both in terms of training and inference, for acquiring new languages. Prior works~\citep{pfeiffer2020madx,ansell2021madg,pfeiffer2022lifting} alleviate the inference inefficiency bottleneck of the curse of multilinguality by introducing modularity in MLMs through language adapters. This modularity allows MLMs to scale the number of parameters with minimal cost on the training and inference speed. One limitation of modular MLMs is that, as shown in Figure~\ref{fig:overview}, the language of the input needs to be known prior to the inference step for selecting the language adapter. Nevertheless, multilingual evaluations of these modular MLMs make an assumption that an ideal language identification is given and use the language metadata provided on the evaluation data to select the correct language adapter. This produces a gap between modular MLMs in the simulated setting and in the real multilingual scenario. In this work, we address the evaluation gap and further discuss how to mitigate the limitation of modular MLMs.

\section{Related Works}

\paragraph{Multilingual Language Model} 
MLMs~\citep{conneau2020xlmr,liu2020mbart,xue2021mt5,bigscience2022bloom} are effective for solving various language understanding and generation in various languages~\citep{hu2020xtreme,wilie2020indonlu,cahyawijaya2021indonlg,adelani2022masakhaner,kumar2022indicnlg}. To solve the curse of multilinguality of MLMs, the modular MLM approach is introduced. MAD-X~\citep{pfeiffer2020madx} and MAD-G~\cite{ansell2021madg} use adapt MLMs to new languages by using language adapters. X-MOD~\citep{pfeiffer2022lifting} introduces modularity during pre-training which better aligns modular MLMs across languages.

\paragraph{Language Identification (LID)} The LID task is introduced over five decades ago~\citep{gold1967lid}. Since then, various methods for LID have been introduced, such as n-gram similarity~\citep{cavnar1994ngrambased}, naive bayes~\citep{baldwin2010lid,lui2012langid,sites2013cld2}, and gaussian mixture~\citep{lui2014polyglot}. More recently, embedding-based methods using character~\citep{salcianu2020cld3} and subwords~\citep{joulin2017bag} have also been introduced. In this work, we explore the effect of utilizing these LID modules on the performance of modular MLMs.

\section{Experimental Setting}

For our experiments, we utilize MASSIVE~\citep{fitzgerald2022massive}, a multilingual intent classification dataset covering 52 typologically-diverse languages. We select 24 languages from MASSIVE and group them into 3 different resource groups based on the language size in CommonCrawl
\footnote{\url{https://commoncrawl.github.io/cc-crawl-statistics/plots/languages}}
, i.e., high-resource languages (HRL), medium-resource languages (MRL), and low-resource languages (LRL). A detailed list of languages under study and the resource grouping is described in Appendix~\ref{app:lang-under-study}. For the LID, we incorporate 5 off-the-shelf LID models, i.e., LangDetect~\citep{shuyo2011langdetect}, langid.py~\citep{lui2012langid}, FastText LID~\citep{joulin2017bag}, CLD2~\citep{sites2013cld2}, and CLD3~\citep{salcianu2020cld3}. We evaluate these LIDs and take the best two LIDs for the multilingual evaluation with unknown languages. For the modular MLM, we utilize MAD-X~\cite{pfeiffer2020madx} with mBERT backbone. We compare the MAD-X with LID against two direct fine-tuned MLMs and MAD-X without LID. We use accuracy score as the evaluation metric in our experiment.

\begin{table}[!t]
\centering
\begin{minipage}{.49\textwidth}
    \centering
    \resizebox{0.82\textwidth}{!}{
        \begin{tabular}{lcccc}
        \toprule
        \multicolumn{1}{c}{\bf LID Model} & \multicolumn{1}{c}{\bf HRL}  & \multicolumn{1}{c}{\bf MRL} & \multicolumn{1}{c}{\bf LRL} & \multicolumn{1}{c}{\bf AVG} \\ 
        \midrule
        \multicolumn{5}{c}{\textit{Fully support languages under study}} \\ 
        \midrule
        FastText & \bf{97.22} & \underline{96.26} & \underline{88.96} & \bf{93.89} \\
        CLD3 & \underline{87.84} & 89.30 & \bf{91.47} & \underline{89.57} \\
        CLD2 & 76.07 & 90.85 & 85.14 & 83.17 \\
        \midrule
        \multicolumn{5}{c}{\textit{Partially support languages under study\footnotemark}} \\
        \midrule
        langid.py & 92.00 & 93.04 & 76.12 & 86.31 \\
        LangDetect & 69.26 & \bf{96.45} & 42.97 & 66.20 \\
        \bottomrule
        \end{tabular}
    }
    \caption{Accuracy score of LIDs on MASSIVE. Most LIDs perform well on \textbf{HRL} and \textbf{MRL}, but the score falls short on \textbf{LRL}. \textbf{Bold} and \underline{underline} denote first and second best, respectively.}
    \label{tab:lid-result}
\end{minipage}
\hfill
\begin{minipage}{.49\textwidth}
    \vspace{-1pt}
    \centering
    \resizebox{0.95\textwidth}{!}{
        \begin{tabular}{lcccc}
        \toprule
        \multicolumn{1}{c}{\bf NLU Model} & \multicolumn{1}{c}{\bf HRL} & \multicolumn{1}{c}{\bf MRL} & \multicolumn{1}{c}{\bf LRL} & \multicolumn{1}{c}{\bf AVG} \\
        \midrule
        \multicolumn{5}{c}{\textit{Direct fine-tuning}} \\ 
        \midrule
        XLMR & \bf{86.03} & \bf{84.76} & \bf{83.20} & \bf{84.65} \\
        mBERT & 84.76 & 82.50 & 80.62 & 82.64 \\
        \midrule
        \multicolumn{5}{c}{\textit{Language adapter tuning}} \\ 
        \midrule
        MAD-X (No LID) & 83.30 & 80.96 & 79.46 & 81.27 \\
        MAD-X (FastText) & 75.21 & 78.08 & 72.46 & 74.90 \\
        MAD-X (CLD3) & 72.90 & 75.20 & 72.89 & 73.47 \\
        \bottomrule
        \end{tabular}
    }
    \caption{Accuracy score of MLMs on MASSIVE. Incorporating LID decays the performance of the language-adapter model. \textbf{Bold} denotes the best performance.}
    \label{tab:mlm-result}
\end{minipage}
\end{table}

\footnotetext{We zero out the performance for all the unsupported languages.}

\section{Result \& Discussion}

Based on the result of the LID experiment in Table~\ref{tab:lid-result}, we select FastText and CLD3 for evaluating modular MLMs with unknown languages. The modular MLMs result is shown in Table~\ref{tab:mlm-result}. For the modular MLM without LID, our result aligns with prior works~\cite{pfeiffer2020madx,ansell2021madg} yielding a slightly lower score compared to the direct fine-tuned models. Both modular MLMs with LID produce an even lower performance in all language resource groups compared to the modular MLM without LID, resulting in a gap of $\sim$7-8\% accuracy score over all language groups. The detailed result of our experiment is shown in Appendix~\ref{app:experiment-result}.

We clearly observe that existing off-the-shelf LID is far from the ideal case which widens the gap to the direct fine-tuning approach and raises an open question for closing the performance gap. To address the question, it is important to understand the limitations of using modular MLMs with off-the-shelf LIDs. Several potential limitations that might occur include: 1) distribution shift of LIDs caused by domain and time differences, 2) label mismatch between LID and the language adapter, and 3) other linguistic problems that affect LIDs such as code-mixing and creole language. 
% Solving these limitations is necessary, but it might not be sufficient to close the performance gap. 
We leave the exploration of the solution to these potential limitations for future works.

\section{Conclusion}

In this work, we show the limitation of modular multilingual language models (MLMs) in inferencing with unknown languages. We evaluate the effect of using off-the-shelf LID modules on the evaluation of modular MLMs. Our result suggests that using off-the-shelf LID modules significantly decreases the performance of modular MLMs by $\sim$7-8\% accuracy which widens the gap between modular MLMs and non-modular MLMs. In addition, we discuss several potential limitations that might contribute to the performance gap of using off-the-shelf LID with modular MLMs.

\subsubsection*{URM Statement}

All authors of this paper qualify as an underrepresented minority (URM) for the ``Tiny Papers'' track at ICLR 2023.

% Please include this URM Statement section at the end of the paper but before the references before. In your anonymized submission, we recommend stating ``The authors acknowledge that at least one key author of this work meets the URM criteria of ICLR 2023 Tiny Papers Track.'' For the camera ready version, we ask authors to identify which author(s) meet the URM criteria, e.g., ``Author TFB meets the URM criteria of ICLR 2023 Tiny Papers Track.'' The authors are also welcome to come up with their own phrases to affirm meeting this criterion.

% first author...
% last author..

% ITB, Indonesia, Bu ayu as supervisor
% Bu ayu female
% not 30 years
% non white, southeast asian

\bibliography{iclr2023_conference_tinypaper}
\bibliographystyle{iclr2023_conference_tinypaper}

\appendix

\section{Language Under Study}
\label{app:lang-under-study}

We provide the list of all languages under study along with the language resource group in Table~\ref{tab:lang-study}. Language resource is grouped by the size of language data in CommonCrawl, i.e., high-resource languages (\textbf{$\geq$1\%}, medium-resource languages ((\textbf{$\geq$0.1\%}), and low-resource languages (\textbf{$<$1\%}).

\begin{table}[ht]
\centering
\resizebox{0.6\linewidth}{!}{
    \begin{tabular}{l|c|c|c}
    \toprule
    \textbf{Language} & \textbf{\#Speaker} & \textbf{CC Size} & \textbf{Resource Group} \\ \midrule
    ar-SA & 360M & 0.665\% & MRL \\
    bn-BD & 300M & 0.093\% & LRL \\
    de-DE & 95M & 5.662\% & HRL \\
    el-GR & 13.5M & 0.597\% & MRL \\
    en-US & 373M & 46.320\% & HRL \\
    es-ES & 493M & 4.435\% & HRL \\
    fi-FI & 5.4M & 0.398\% & LRL \\
    fr-FR & 300M & 4.604\% & HRL \\
    hi-IN & 528M & 0.155\% & LRL \\
    hu-HU & 13M & 0.599\% & MRL \\
    hy-AM & 5.4M & 0.032\% & LRL \\
    id-ID & 300M & 0.781\% & MRL \\
    is-IS & 0.3M & 0.038\% & LRL \\
    ja-JP & 128M & 4.532\% & HRL \\
    jv-ID & 82M & 0.002\% & LRL \\
    ka-GE & 3.7M & 0.037\% & LRL \\
    ko-KR & 79.3M & 0.679\% & MRL \\
    lv-LV & 1.2M & 0.082\% & LRL \\
    my-MM & 33M & 0.012\% & LRL \\
    pt-PT & 250M & 1.482\% & HRL \\
    ru-RU & 258M & 5.717\% & HRL \\
    vi-VN & 70M & 0.962\% & MRL \\
    zh-CN & 920M & 4.837\% & HRL \\
    zh-TW & 4.6M & 4.837\%\footnotemark & HRL \\
    \bottomrule
    \end{tabular}
}
\caption{List of languages under study in our experiments. The number of speaker information is retrieved from Wikipedia.}
\label{tab:lang-study}
\end{table}

\footnotetext{We use the same number of zh-CN and zh-TW, since there is no Chinese (zh) language variation in CommonCrawl.}

\section{Detailed Experiment Result}
\label{app:experiment-result}
We provide the complete  per language result for the language identification and the modular MLMs experiments in Table~\ref{tab:lid-complete} and Table~\ref{tab:mlm-complete}.

\begin{table}[ht]
\centering
\resizebox{0.66\linewidth}{!}{
    \begin{tabular}{l|c|c|c|c|c}
    \toprule
    \textbf{Language} & \textbf{LID-Fasttext} & \textbf{CLD3} & \textbf{CLD2} & \textbf{langid} & \textbf{LangDetect} \\
    \midrule
    ar-SA & 94.25 & 86.45 & 81.58 & 91.78 & 94.13 \\
    bn-BD & 99.72 & 97.52 & 89.57 & 96.93 & 99.76 \\
    de-DE & 97.70 & 88.59 & 89.73 & 92.83 & 82.54 \\
    el-GR & 99.68 & 96.91 & 99.77 & 99.84 & 99.64 \\
    en-US & 98.61 & 79.44 & 93.43 & 93.96 & 87.82 \\
    es-ES & 96.20 & 78.24 & 73.14 & 86.87 & 86.55 \\
    fi-FI & 97.70 & 92.91 & 92.90 & 92.08 & 96.09 \\
    fr-FR & 98.35 & 87.53 & 85.23 & 94.77 & 94.80 \\
    hi-IN & 98.44 & 88.21 & 97.83 & 87.94 & 93.54 \\
    hu-HU & 98.54 & 92.24 & 93.89 & 95.34 & 96.71 \\
    hy-AM & 99.90 & 98.37 & 99.92 & 99.17 & 0.00 \\
    id-ID & 87.20 & 65.86 & 73.54 & 72.68 & 89.32 \\
    is-IS & 89.93 & 92.64 & 90.88 & 92.97 & 0.00 \\
    ja-JP & 99.41 & 96.63 & 99.04 & 99.11 & 96.23 \\
    jv-ID & 24.75 & 68.10 & 0.00 & 22.04 & 0.00 \\
    ka-GE & 99.56 & 98.49 & 99.95 & 99.65 & 0.00 \\
    ko-KR & 99.50 & 98.47 & 99.03 & 99.96 & 99.36 \\
    lv-LV & 90.73 & 90.06 & 95.25 & 94.33 & 97.32 \\
    my-MM & 99.93 & 96.90 & 99.97 & 0.00 & 0.00 \\
    pt-PT & 92.17 & 83.42 & 77.39 & 77.74 & 84.05 \\
    ru-RU & 99.27 & 84.48 & 82.35 & 83.79 & 91.32 \\
    vi-VN & 98.41 & 95.85 & 97.26 & 98.62 & 99.53 \\
    zh-CN & 97.55 & 98.07 & 84.33 & 99.64 & 0.00 \\
    zh-TW & 95.76 & 94.19 & 0.03 & 99.31 & 0.00 \\ \midrule
    \textbf{Average} & \textbf{93.89} & \textbf{89.57} & \textbf{83.17} & \textbf{86.31} & \textbf{66.20} \\
    \bottomrule
    \end{tabular}
}
\caption{Per language results of language identification evaluation in MASSIVE.}
\label{tab:lid-complete}
\end{table}

\begin{table}[ht]
\centering
\resizebox{0.66\linewidth}{!}{
    \begin{tabular}{l|c|c|c|c|c}
    \toprule
    \textbf{Language} & \textbf{XLMR} & \textbf{mBERT} & \textbf{MAD-X} & \textbf{\begin{tabular}[c]{@{}c@{}}MAD-X\\w/ FastText\end{tabular}} & \textbf{\begin{tabular}[c]{@{}c@{}}MAD-X\\w/ CLD3\end{tabular}} \\
    \midrule
    ar-SA & 79.32 & 78.35 & 75.72 & 71.92 & 67.79 \\
    bn-BD & 83.25 & 80.23 & 78.61 & 76.36 & 74.95 \\
    de-DE & 85.54 & 83.59 & 81.81 & 79.49 & 76.90 \\
    el-GR & 85.07 & 81.74 & 80.93 & 79.56 & 78.51 \\
    en-US & 88.16 & 86.45 & 85.78 & 83.89 & 83.15 \\
    es-ES & 86.18 & 84.97 & 82.58 & 80.97 & 76.43 \\
    fi-FI & 85.24 & 82.55 & 82.55 & 79.86 & 77.07 \\
    fr-FR & 86.48 & 86.11 & 83.69 & 82.35 & 80.03 \\
    hi-IN & 84.63 & 82.38 & 80.73 & 78.14 & 72.73 \\
    hu-HU & 85.68 & 82.65 & 81.57 & 80.13 & 76.40 \\
    hy-AM & 84.23 & 81.20 & 80.43 & 78.78 & 77.91 \\
    id-ID & 86.52 & 84.67 & 82.01 & 76.03 & 69.30 \\
    is-IS & 84.16 & 82.21 & 80.40 & 71.49 & 73.57 \\
    ja-JP & 85.78 & 84.70 & 83.22 & 82.04 & 81.27 \\
    jv-ID & 81.20 & 81.57 & 78.58 & 45.70 & 59.68 \\
    ka-GE & 79.19 & 75.25 & 73.23 & 70.85 & 70.17 \\
    ko-KR & 85.51 & 84.30 & 82.99 & 81.14 & 80.56 \\
    lv-LV & 84.73 & 82.18 & 82.08 & 74.58 & 74.95 \\
    my-MM & 82.18 & 78.01 & 78.48 & 76.36 & 74.98 \\
    pt-PT & 86.35 & 85.27 & 83.59 & 80.56 & 77.77 \\
    ru-RU & 86.65 & 83.96 & 83.52 & 81.74 & 75.45 \\
    vi-VN & 86.48 & 83.32 & 82.52 & 79.72 & 78.61 \\
    zh-CN & 85.41 & 85.24 & 84.23 & 53.09 & 52.69 \\
    zh-TW & 83.73 & 82.55 & 81.27 & 52.79 & 52.45 \\ \midrule
    \textbf{Average} & \textbf{84.65} & \textbf{82.64} & \textbf{81.27} & \textbf{74.90} & \textbf{73.47} \\
    \bottomrule
    \end{tabular}
}
\caption{Per language accuracy score of multilingual language models in MASSIVE.}
\label{tab:mlm-complete}
\end{table}

\end{document}